\documentclass[pdflatex,sn-mathphys-num]{sn-jnl}

\usepackage{graphicx}%
\usepackage{multirow}%
\usepackage{amsmath,amssymb,amsfonts}%
\usepackage{amsthm}%
\usepackage{mathrsfs}%
\usepackage[title]{appendix}%
\usepackage{xcolor}%
\usepackage{textcomp}%
\usepackage{manyfoot}%
\usepackage{booktabs}%
\usepackage{algorithm}%
\usepackage{algorithmicx}%
\usepackage{algpseudocode}%
\usepackage{listings}%
\usepackage{algorithm}
\usepackage{algpseudocode}
\usepackage{amsmath}

\theoremstyle{thmstyleone}%
\theoremstyle{thmstyletwo}%

\theoremstyle{thmstylethree}%

\begin{document}

\title[Article Title]{Randomized Spline Trees for Functional Data Classification: Theory and Application to Environmental Time Series}


\author[1]{\fnm{Donato} \sur{Riccio}}\email{donato.riccio@studenti.unicampania.it}

\author*[2]{\fnm{Fabrizio} \sur{Maturo}}\email{fabrizio.maturo@unimercatorum.it}

\author[3]{\fnm{Elvira} \sur{Romano}}\email{elvira.romano@unicampania.it}

\affil[1]{Machine Learning Engineer and Student in the Data Science Master's Degree Program of the University of Campania Luigi Vanvitelli, Caserta, Italy}

\affil*[2]{Department of Economics, Statistics and Business, Faculty of Technological and Innovation Sciences, Universitas Mercatorum, Rome, Italy}

\affil[3]{Department of Mathematics and Physics, University of Campania Luigi Vanvitelli, Caserta, Italy}


\abstract{Functional data analysis (FDA) and ensemble learning can be powerful tools for analyzing complex environmental time series. Recent literature has highlighted the key role of diversity in enhancing accuracy and reducing variance in ensemble methods.This paper introduces Randomized Spline Trees (RST), a novel algorithm that bridges these two approaches by incorporating randomized functional representations into the Random Forest framework. RST generates diverse functional representations of input data using randomized B-spline parameters, creating an ensemble of decision trees trained on these varied representations. We provide a theoretical analysis of how this functional diversity contributes to reducing generalization error and present empirical evaluations on six environmental time series classification tasks from the UCR Time Series Archive. Results show that RST variants outperform standard Random Forests and Gradient Boosting on most datasets, improving classification accuracy by up to 14\%. The success of RST demonstrates the potential of adaptive functional representations in capturing complex temporal patterns in environmental data. This work contributes to the growing field of machine learning techniques focused on functional data and opens new avenues for research in environmental time series analysis.}

\keywords{FDA, Ensemble Learning, Diversity, Time Series Classification}



\maketitle

\section{Introduction}
Technological advancements have revolutionized the collection of massive volumes of environmental data. Today, various instruments and sensors can record different types of signals to monitor environmental conditions, predict climate patterns, and assess ecosystem health. For instance, satellite imagery helps track land use changes and deforestation, while weather stations provide continuous data on temperature, precipitation, and atmospheric conditions. Supervised and unsupervised classification, feature extraction, and dimensionality reduction techniques are among the most used strategies for dealing with these data types. 

Learning from high-dimensional environmental data is always challenging due to the curse of dimensionality that may lead to data sparsity, difficulties in selecting a unique statistical model, multicollinearity, and distance concentration \cite{verleysen2005curse}.  For these reasons, in recent decades, research on the classification of big data coming from environmental domains and dimensionality reduction techniques have assumed a fundamental role in many areas, such as ecology, climatology, remote sensing, environmental science, and data science \cite{Runting2020}. Functional Data Analysis (FDA) has emerged as a robust framework for analyzing and modeling data that are functional in nature \cite{Ramsay2005, ferraty2006nonparametric}. Environmental signals  observed over time or space can be interpreted as functions in their respective domains and, consequently, be treated as single objects . 

FDA has been applied to various domains in climatology and environmental studies. For instance, Ghumman et al. utilized FDA techniques to analyze models predicting temperature and precipitation under different climate scenarios \cite{ghumman2020}. Similarly, Holtanová et al. effectively employed FDA methodologies to distinguish time series data in climatology \cite{holtanova2019}.  Burdejová and Härdle proposed a dynamic functional factors model to enhance the analysis of climatological data such as temperature and precipitation \cite{burdejova2019}. Additionally, Villani, Romano and Mateu introduce a new FDA-based methodology for selecting and validating climate models, with a focus on predicting average temperatures in the Campania region (Italy) \cite{Villani2024}.

The application of machine learning (ML) to FDA  has evolved significantly over the past few decades. Initially, FDA was primarily dominated by classical statistical methods, such as Functional Principal Component Analysis (FPCA) and Functional Linear Models (FLM), which laid the groundwork for representing and interpreting functional data \cite{ramsay2004functional, ferraty2006nonparametric}. These techniques focused on capturing the inherent variability and structure in functional data, which can be represented as curves or functions over a continuum, such as time or space. The integration of ML techniques into FDA began to gain momentum with the rise of more advanced computational methods. Early approaches leveraged support vector machines (SVMs) and neural networks (NNs)  to handle the complexities of functional data. For instance, Rossi and Conan-Guez  \cite{Rossi_2006} demonstrated the potential of SVMs for functional classification, highlighting their ability to manage the high dimensionality and smooth nature of functional data. Similarly, Multi-layer Perceptrons were explored for their flexibility in modeling non-linear relationships in functional data \cite{rossi2002functional}.

There has been a growing interest in ensemble learning techniques for FDA recently. Ensemble learning, which combines multiple models to improve predictive performance, has shown remarkable success in various domains of ML \cite{diettichensemble}. 
These techniques have shown great promise in addressing the challenges of functional data classification tasks.
\cite{Gregorutti_2015,maturo2022pooling, maturo2024combining}.
Its application to FDA is particularly promising due to the complex nature of functional data, which often benefits from the diverse perspectives provided by multiple models.

The intersection of ensemble methods and FDA presents a rich area for innovation. Maturo et al. introduced the Functional Random Forest (FRF) \cite{maturo2022pooling}, which effectively applied Random Forest (RF) principles to functional data using B-spline coefficients.  
More recently, Riccio et al. presented the Functional Voting Classifier (FVC) \cite{riccio2024supervisedlearningensemblesdiverse}, a novel ensemble architecture designed for functional data classification. By training multiple base models on different functional representations of the input data and aggregating their predictions through voting, the FVC demonstrated significant improvements in accuracy and robustness compared to individual models across various real-world datasets. A key finding of this work was that inducing diversity through different functional representations leads to increased classification performance.

We present a new framework called Randomized Spline Trees (RST), motivated by the need for an ensemble algorithm specifically tailored to functional inputs. Previous research has demonstrated the benefits of diversity in such algorithms, and our aim was to integrate these insights into a novel model. RST is designed to overcome the limitations of previous approaches such as FVC and FRF, particularly addressing the constraint of having a fixed basis representation for every ensemble member. Building upon these insights,  RST represents an evolution of the Random Forest algorithm tailored for functional data inputs. The core innovation of RST lies in its approach to inducing diversity among the constituent trees. While standard Random Forests achieve diversity primarily through bagging and random feature subset selection, RST introduces an additional layer of randomization in the functional representation.
By fully randomizing the basis representations, RST aims to overcome the fixed basis representations of FVC and FRF.

Specifically, each tree in an RST ensemble is trained on a unique functional representation of the input data obtained through B-spline basis expansions with randomized parameters. By varying the number of basis functions $K$ and the order of the splines $o$ for each tree, RST creates a highly diverse set of perspectives on the underlying functional data. The parameter randomization creates a set of weak learners, which are preferred in ensemble learning \cite{freund1997decision}. This approach aims to capture a wide variety of features and patterns that may be present at different scales or levels of smoothness in the input functions. By combining both model-level and representation-level randomization, RST aims to create ensembles with greater diversity than standard Random Forests or even the FVC approach.

This paper makes several key contributions. First, we introduce the RST algorithm, a novel extension of Random Forests for functional data classification. In addition, we provide a theoretical analysis of how randomized functional representations contribute to ensemble diversity and relate this to existing work on the bias-variance tradeoff in Random Forests. Lastly, we present extensive empirical evaluations comparing RST to standard Random Forests and other machine learning algorithms across a diverse set of functional classification tasks, demonstrating its superior accuracy.

The remainder of this paper is organized as follows: Section \ref{tf} presents the theoretical framework underpinning our approach, including the concepts of ensemble diversity, functional diversity through B-spline randomization, the functional coefficients matrix, and a generalization error analysis. Section \ref{rst} introduces the RST algorithm, detailing its structure and implementation. Section \ref{appl} describes our experimental evaluation, including the datasets used, experimental setup, results, and performance analysis. 

\section{Theoretical Framework} \label{tf}
\subsection{Diversity in ensemble learning}

Diversity is fundamental in ensemble learning, significantly enhancing performance and robustness. A diverse ensemble can correct individual model errors, leading to better overall accuracy \cite{kuncheva2003measures}. Dietterich \cite{diettichensemble} identified three key reasons for the superiority of ensemble methods over single classifiers. With limited training data, multiple hypotheses may perform equally well on the training set. An ensemble can average these hypotheses to reduce the risk of choosing the wrong hypothesis (\textit{statistical reason}). Many learning algorithms involve local searches that may get stuck in local optima. An ensemble constructed by running the local search from different starting points may better approximate the true unknown function \textit{(computational reason)}.
In many cases, the true function cannot be represented by any of the hypotheses in the hypothesis space. By forming weighted sums of hypotheses, it may be possible to expand the space of representable functions (\textit{representational reason}).

Diverse classifiers can reduce the risk of choosing the wrong hypothesis, avoid local optima, and expand the representable function space.
Diversity in ensemble learning is critical because it allows the ensemble to capture a broader range of patterns and make more robust predictions. Kuncheva and Whitaker \cite{kuncheva2003measures} discussed the concepts of good and bad diversity, emphasizing that not all diversity is beneficial. Good diversity refers to variations among models that contribute positively to the ensemble's performance, while bad diversity introduces variations that do not improve and may even degrade performance.

Diversity in ensemble learning can be categorized into \textit{data-based} and \textit{model-based diversity} \cite{Gong_2019}. Data-based diversity is achieved by varying the training data for each model in the ensemble. Common techniques include \textit{bagging}, which trains models on random subsets of the training data, sampled with replacement \cite{breiman1996bagging} and \textit{boosting}, which reweights training instances to focus on those misclassified by previous models \cite{freund1999short}. These methods help models capture different aspects of the input space, providing complementary information.

\textit{Model-based} diversity involves varying the models themselves, even with the same training data. It can be achieved by using various learning algorithms like decision trees, neural networks, and SVMs \cite{Gong_2019}; training models with different hyperparameters, such as the depth of decision trees or the number of hidden layers in neural networks; randomly initializing weights in methods like neural networks; using dropout, a technique in neural networks where units are randomly dropped during training to prevent overfitting and create an ensemble of networks \cite{srivastava2014dropout}.

Recent research has identified a third category: \textit{functional-based diversity}. 
This approach, introduced by Riccio et al. \cite{riccio2024supervisedlearningensemblesdiverse}, explores novel ways to induce diversity by introducing variations in the functional representations of input data. This emerging category offers new possibilities for enhancing ensemble performance, particularly in domains where data is inherently functional or can be represented as such.

\subsection{Functional diversity through B-spline randomization}

B-splines form a flexible basis for approximating curves. Let us define a knot sequence $\boldsymbol{\xi} = (\xi_1, \ldots, \xi_{K+o+1})$ where $K$ is the number of basis functions and $o$ is the order (degree + 1) of the B-spline. The $i$-th B-spline basis function of order $o$, denoted $B_{i,o}(t)$, is defined recursively using the Cox-de Boor formula \cite{de1972calculating}:

\begin{equation}
B_{i,1}(t) = 
\begin{cases}
1 & \text{if } \xi_i \leq t < \xi_{i+1} \\
0 & \text{otherwise}
\end{cases}
\end{equation}

\begin{equation}
B_{i,o}(t) = \frac{t - \xi_i}{\xi_{i+o-1} - \xi_i} B_{i,o-1}(t) + \frac{\xi_{i+o} - t}{\xi_{i+o} - \xi_{i+1}} B_{i+1,o-1}(t)
\end{equation}

A curve $x(t)$ can then be approximated as a linear combination of these basis functions:

\begin{equation}
x(t) \approx \hat{x}(t) = \sum_{i=1}^K c_i B_{i,o}(t)
\end{equation}

Where $c_i$ is the $i$th spline coefficient.

The order $o$ of a B-spline determines its degree and continuity properties. A B-spline of order $o$ is a piecewise polynomial of degree $o-1$ with continuous derivatives up to order $o-2$ at the knots. 

Consider two B-spline approximations of the same curve $x(t)$ with different orders $o_1$ and $o_2$, but the same number of basis functions $K$:

\begin{equation}
\hat{x}_{o_1}(t) = \sum_{i=1}^K c_i^{(o_1)} B_{i,o_1}(t)
\end{equation}

\begin{equation}
\hat{x}_{o_2}(t) = \sum_{i=1}^K c_i^{(o_2)} B_{i,o_2}(t)
\end{equation}

\begin{figure}
    \centering
    \includegraphics[width=1\linewidth]{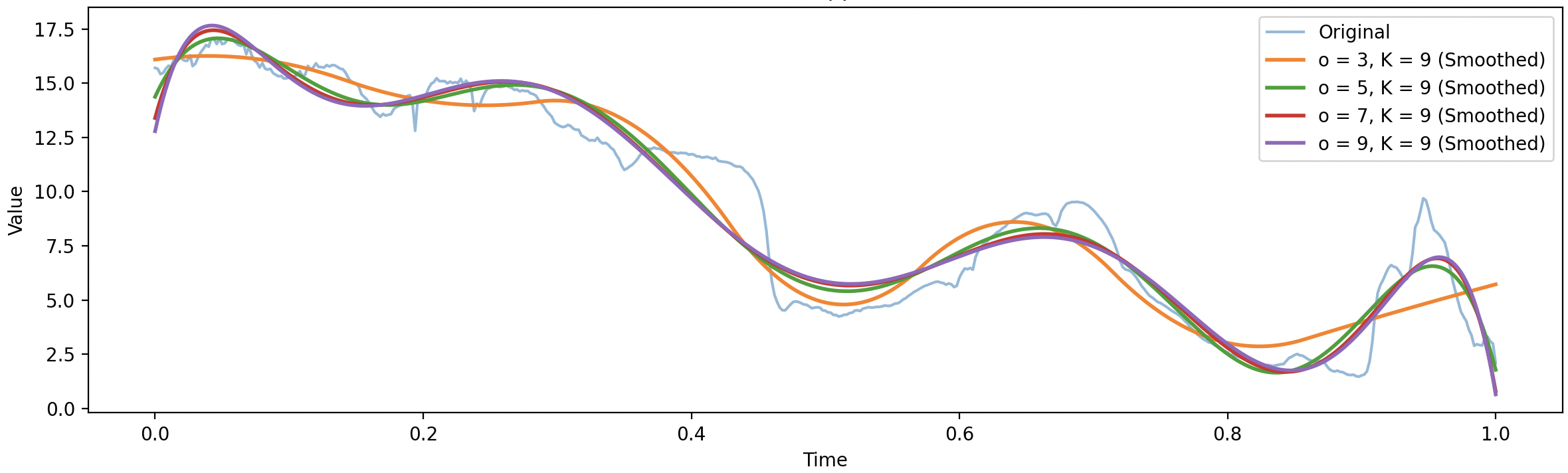}
    \caption{Functional representations of the first instance in the Rock dataset, with fixed $K$ and varying $o$.}
    \label{fig:fixed_k}
\end{figure}

Figure \ref{fig:fixed_k} shows functional representations with a fixed number of basis functions in increasing order. The critical difference lies in the support and smoothness of the basis functions. Higher-order B-splines have wider support and are smoother, which affects their ability to capture local features.

Let us consider the effect of varying the number of basis functions $K$, while keeping the order $o$ fixed. We have two approximations:
\begin{equation}
\hat{x}_{K_1}(t) = \sum_{i=1}^{K_1} c_i^{(K_1)} B_{i,o}(t)
\end{equation}
\begin{equation}
\hat{x}_{K_2}(t) = \sum_{i=1}^{K_2} c_i^{(K_2)} B_{i,o}(t)
\end{equation}

\begin{figure}
    \centering
    \includegraphics[width=1\linewidth]{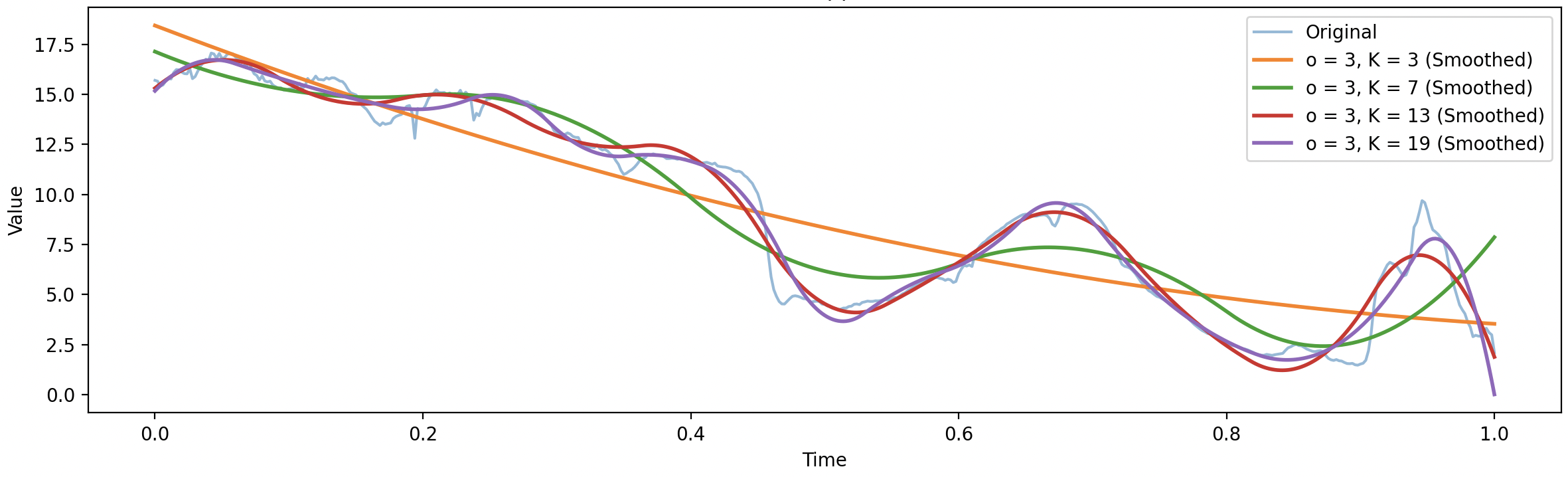}
    \caption{Functional representations of the first instance in the Rock dataset, with fixed $o$ and varying $K$.}
    \label{fig:fixed_o}
\end{figure}

Figure \ref{fig:fixed_o} shows different functional representations with an increasing number of basis functions. Increasing $K$ allows for more localized control over the curve shape, effectively increasing the number of knots. This strategy can lead to better approximation of fine details but may also increase the risk of overfitting.

The core of RST's diversity-inducing mechanism lies in the randomization of $o$ and $K$ for each tree in the ensemble. Let $T$ be the number of trees in the RST ensemble. For each tree $t = 1,\ldots,T$, we randomly select:
\begin{align}
o_t &\sim \text{Uniform}\{o_{\min}, \ldots, o_{\max}\} \\
K_t &\sim \text{Uniform}\{K_{\min}, \ldots, K_{\max}\}
\end{align}
where $o_{\min}, o_{\max}, K_{\min},$ and $K_{\max}$ are hyperparameters of the RST algorithm.

This randomization leads to $T$ different functional representations of the input data. We can define a distance measure between two B-spline approximations to quantify the diversity between these representations.

Let $\hat{x}_i^{(s)}(t)$ and $\hat{x}_i^{(r)}(t)$ be two different B-spline approximations of $x_i(t)$, corresponding to trees $s$ and $r$. We can define the $L^2$ distance between these approximations as \cite{Hastie2009}:
\begin{equation}
d(\hat{x}_i^{(s)}, \hat{x}_i^{(r)}) = \left(\int_a^b (\hat{x}_i^{(s)}(t) - \hat{x}_i^{(r)}(t))^2 dt\right)^{1/2}
\end{equation}

The pairwise average diversity of an ensemble of T trees is given by:
\begin{equation}
\text{D} = \frac{2}{T(T-1)} \sum_{1 \leq s < r \leq T} d(\hat{x}_i^{(s)}, \hat{x}_i^{(r)})
\end{equation}

This diversity measure quantifies the functional representations' differences, a key feature of the RST algorithm. By randomizing the functional parameters, RST leads to an increased $D$.

This measure provides a comprehensive view of the diversity of functional representations across the entire ensemble. A higher value of $F$ indicates greater functional diversity, which is a form of \textit{data-based diversity} desirable property in ensemble methods as it can lead to improved generalization. \cite{kuncheva2003measures}

The randomization of $o$ and $K$ in RST directly contributes to increasing this functional diversity, as each tree in the ensemble is likely to have a unique combination of order and number of basis functions, leading to distinct functional representations of the input data.

While the pairwise $L_2$ distance provides a useful measure of diversity between functional representations, we can extend our analysis to include additional measures that capture different aspects of ensemble diversity. These measures offer complementary perspectives on how the randomized functional representations in RST contribute to the overall diversity of the ensemble. Analogous to the Gini index used in decision tree splitting, we can define a quadratic difference measure for functional representations. This measure emphasizes larger differences between representations, potentially capturing more significant structural variations.
For a pair of functional representations $\hat{x}_i^{(s)}(t)$ and $\hat{x}_i^{(r)}(t)$, we define the quadratic difference as:

\begin{equation}
    Q(\hat{x}_i^{(s)}, \hat{x}_i^{(r)}) = \int_a^b (\hat{x}_i^{(s)}(t) - \hat{x}_i^{(r)}(t))^2 dt.
\end{equation}

The ensemble quadratic diversity measure can then be defined as:

\begin{equation}
Q_D = \frac{2}{T(T-1)} \sum_{1 \leq s < r \leq T} Q(\hat{x}_i^{(s)}, \hat{x}_i^{(r)}).
\end{equation}

This measure gives more weight to pairs of representations that differ significantly, potentially identifying subsets of the ensemble that capture substantially different aspects of the underlying functional data.
While pairwise measures provide insights into the diversity between individual trees, we can also consider the overall variance of the functional representations across the ensemble. This approach treats the set of functional representations as a distribution and measures its spread.
Let $\bar{x}_i(t)$ be the mean functional representation across the ensemble:

\begin{equation}
\bar{x}_i(t) = \frac{1}{T} \sum_{s=1}^T \hat{x}_i^{(s)}(t).
\end{equation}

We can then define the functional variance measure as:

\begin{equation}
V_F = \frac{1}{T} \sum_{s=1}^T \int_a^b (\hat{x}_i^{(s)}(t) - \bar{x}_i(t))^2 dt.
\end{equation}

This quantity measures how much, on average, individual functional representations deviate from the ensemble mean representation. A higher $V_F$ indicates greater diversity in how different trees in the ensemble represent the underlying functional data.

\subsection{The functional coefficients matrix}
At the base of the RST algorithm lies the concept of the functional coefficients matrix, denoted as $\mathbf{C}_{o,K}$. This matrix transformation converts the original functional data into a format suitable for decision tree learning while preserving the functional nature of the data.

For a given B-spline basis defined by order $o$ and number of basis functions $K$, the functional coefficients matrix $\mathbf{C}_{o,K}$ is an $N \times K$ matrix, where $N$ is the number of observations in the dataset. Each row of $\mathbf{C}_{o,K}$ corresponds to an observation, and each column corresponds to a coefficient of the B-spline basis expansion.

We can express $\mathbf{C}_{o,K}$ as:

\begin{equation}
\mathbf{C}_{o,K} = 
\begin{bmatrix}
c_{11} & c_{12} & \cdots & c_{1K} \\
c_{21} & c_{22} & \cdots & c_{2K} \\
\vdots & \vdots & \ddots & \vdots \\
c_{N1} & c_{N2} & \cdots & c_{NK}
\end{bmatrix}
\end{equation}

where $c_{ij}$ is the $j$-th B-spline coefficient for the $i$-th observation.

For each observation $\mathbf{x}_i = (x_{i1}, \ldots, x_{iP})$, the corresponding row in $\mathbf{C}_{o,K}$ is computed by solving the least squares problem:

\begin{equation}
\mathbf{c}_i = (c_{i1}, \ldots, c_{iK}) = \arg\min_{\mathbf{c}} \sum_{p=1}^P \left(x_{ip} - \sum_{j=1}^K c_j B_{j,o}\left(\frac{p}{P}\right)\right)^2
\end{equation}

where $B_{j,o}(t)$ are the B-spline basis functions of order $o$.

The functional coefficients matrix $\mathbf{C}_{o,K}$ transforms the original $P$-dimensional functional data into a $K$-dimensional representation, where typically $K < P$. This mechanism can help in managing high-dimensional functional data.
Despite the dimensionality reduction, $\mathbf{C}_{o,K}$ preserves the essential functional characteristics of the data, as it captures the shape and features of the original functions through the B-spline coefficients.
By using B-spline bases, the representation can adapt to the functions' global and local features, depending on the chosen order $o$ and number of basis functions $K$.
    
In RST, we generate a different $\mathbf{C}_{o_t,K_t}$ for each tree $t$ in the ensemble by randomly varying $o_t$ and $K_t$, introducing diversity in the functional representations.

In the RST algorithm, we generate $T$ different functional coefficient matrices $\{\mathbf{C}_{o_1,K_1}, \ldots, \mathbf{C}_{o_T,K_T}\}$, one for each tree in the ensemble. Each $\mathbf{C}_{o_t,K_t}$ is computed using a different randomly chosen B-spline basis (with parameters $o_t$ and $K_t$).

When training the $t$-th tree, we use $\mathbf{C}_{o_t,K_t}$ as the input data instead of the original functional data. Each decision in the tree is made based on the B-spline coefficients rather than the original function values.
For prediction, a new functional observation is first transformed into its B-spline coefficient representation using each of the $T$ different bases, resulting in $T$ different representations $\mathbf{c}_{new,o_t,K_t}$ for $t = 1, \ldots, T$. These representations are then passed through their respective trees.
By using these diverse functional representations, RST aims to capture different aspects of the functional data, potentially leading to improved predictive performance compared to methods that use a single fixed representation.

\subsection{Practical performance and reliability}

To understand the theoretical foundations of RST, we analyze its generalization error, adapting Breiman's formulation for Random Forests \cite{breiman2001random} to account for the randomized functional representations in RST.

Let us consider an RST ensemble with $T$ trees. For each tree $t$, the input data is represented by the functional coefficient matrix $\mathbf{C}_{o_t,K_t}$, obtained through a B-spline basis expansion with randomly chosen parameters $\theta_t = (o_t, K_t)$, where $o_t$ is the B-spline order and $K_t$ is the number of basis functions.

Let $X, Y$ be random variables representing an input-output pair, and $h(\mathbf{C}_{o,K}(X))$ be a single tree in the ensemble, where $\mathbf{C}_{o,K}(X)$ denotes the functional coefficient matrix for input $X$ using B-spline parameters $(o,K)$.

The RST ensemble prediction is given by:

\begin{equation}
f_{\text{RST}}(X) = \mathbb{E}_{\theta}[h(\mathbf{C}_{o,K}(X))]
\end{equation}

where $\mathbb{E}_{\theta}$ denotes expectation over the random parameters $\theta = (o,K)$.

The generalization error of the RST ensemble can be expressed as:

\begin{equation}
\text{GErr}(X, Y) = \mathbb{E}_{X,Y}[(Y - f_{\text{RST}}(X))^2]
\end{equation}

Following Breiman's analysis, we can decompose this error as:

\begin{equation}
\text{GErr}(X, Y) = \rho(\mathbb{E}_{\theta}[\sigma^2(\mathbf{C}_{o,K}(X))]) + \mathbb{E}_X[(\mathbb{E}_{\theta}[h(\mathbf{C}_{o,K}(X))] - f(X))^2]
\end{equation}

where:
\begin{itemize}
\item $\rho$ is the weighted correlation between the residuals $Y - h(\mathbf{C}_{o,K}(X))$ and $Y - h(\mathbf{C}_{o',K'}(X))$ for $(o,K) \neq (o',K')$
\item $\sigma^2(\mathbf{C}_{o,K}(X)) = \text{Var}_Y[Y | \mathbf{C}_{o,K}(X)]$
\item $f(X) = \mathbb{E}[Y | X]$ is the true regression function
\end{itemize}

The first term, $\rho(\mathbb{E}_{\theta}[\sigma^2(\mathbf{C}_{o,K}(X))])$, represents the average variance of the individual trees, weighted by their correlation. The second term is the squared bias of the ensemble.

In RST, the correlation $\rho$ plays a key role. It captures the correlation induced by the different functional representations. By randomizing the functional representations through different $\mathbf{C}_{o,K}$ matrices, RST aims to reduce the overall correlation $\rho$ compared to standard Random Forests. 

To further analyze the impact of functional representation diversity, we can consider the variance of the RST ensemble:

\begin{equation}
\text{Var}[f_{\text{RST}}(X)] = \rho(\sigma^2(\mathbf{C}_{o,K}(X)))
\end{equation}

where $\sigma^2(\mathbf{C}_{o,K}(X))$ is the average variance of individual trees.

This formulation highlights how reducing $\rho$ through diverse functional representations (i.e., different $\mathbf{C}_{o,K}$ matrices) can lead to a reduction in the overall ensemble variance.

The bias term in RST can be expressed as:

\begin{equation}
\text{Bias}^2 = \mathbb{E}_X[(\mathbb{E}_{\theta}[h(\mathbf{C}_{o,K}(X))] - f(X))^2]
\end{equation}

While individual trees may have a high bias due to their randomized representations (i.e., specific $\mathbf{C}_{o,K}$ matrices), the ensemble can potentially reduce this bias by combining diverse perspectives on the data.

The effectiveness of RST in reducing generalization error depends on achieving a favorable trade-off between reducing correlation (and thus variance) through diverse functional representations and maintaining acceptably low bias. The choice of randomization influences this trade-off ranges for $o$ and $K$, which determine the set of possible $\mathbf{C}_{o,K}$ matrices:

If the ranges for $o$ and $K$ are too narrow, the diversity of $\mathbf{C}_{o,K}$ matrices may be limited, leading to higher correlation and potentially higher generalization error.
Conversely, if the ranges are too broad, individual trees may become too biased due to extreme $\mathbf{C}_{o,K}$ matrices, potentially increasing the overall ensemble bias.

In practice, the optimal ranges for $o$ and $K$ will depend on the specific characteristics of the analyzed functional data. Cross-validation or other model selection techniques can be used to tune these hyperparameters for a given problem.

\section{The Randomized Spline Trees Algorithm} \label{rst}
RST extends the Random Forest algorithm to handle functional data by introducing a novel approach to data representation. This section provides a detailed description of the RST algorithm, highlighting its unique features and how it integrates with the standard Random Forest framework.

Let $\mathcal{D} = \{(\mathbf{x}_i, y_i)\}_{i=1}^N$ be a dataset of $N$ observations, where $\mathbf{x}_i = (x_{i1}, \ldots, x_{iP})$ is a $P$-dimensional vector of predictor variables for the $i$-th observation, and $y_i \in \{1, \ldots, Z\}$ is the corresponding class label. RST constructs an ensemble of $T$ trees, each built on a unique functional representation of the input data.

The key hyperparameters of the RST algorithm are:

\begin{enumerate}
    \item $n\_estimators$: The number of trees in the forest ($T$).
    \item $o_{min}, o_{max}$: The minimum and maximum orders for B-spline bases.
    \item $K_{min}, K_{max}$: The minimum and maximum numbers of basis functions.
    \item $split$: The strategy for selecting split points (\texttt{best} or \texttt{random}).
    \item $bootstrap$: Whether to use bootstrap sampling for each tree (\texttt{True} or \texttt{False}).
\end{enumerate}

The first step is obtaining the functional data representations. For each tree $t = 1, \ldots, T$:

\begin{enumerate}
    \item Construct the B-spline basis functions $B_{j,o_t}(x)$, $j = 1, \ldots, K_t$ using the recursive definition.

    \item Compute the functional coefficient matrix $\mathbf{C}_{o_t,K_t}$ by solving the least squares problem for each observation $\mathbf{x}_i$.
\end{enumerate}

The result of this step is a set of $T$ different functional coefficient matrices $\{\mathbf{C}_{o_1,K_1}, \ldots, \mathbf{C}_{o_T,K_T}\}$, each representing a unique functional representation of the original dataset.

Once the functional representations are obtained, RST leverages the standard Random Forest algorithm to build the ensemble:

\begin{enumerate}
    \item For each tree $t = 1, \ldots, T$:
    \begin{enumerate}
        \item Use $\mathbf{C}_{o_t,K_t}$ as the input data for training the tree.
        \item If $bootstrap = True$, perform bootstrap sampling on the rows of $\mathbf{C}_{o_t,K_t}$.
        \item Grow the tree using the specified $split$ strategy (\textit{best} or \textit{random}) on the columns of $\mathbf{C}_{o_t,K_t}$.
        \item Continue until stopping criteria are met (e.g., maximum depth, minimum samples for split).
    \end{enumerate}
\end{enumerate}

The result is an ensemble of $T$ decision trees, each built on a different functional representation of the input data.

To make a prediction for a new observation $\mathbf{x}_{new}$:

\begin{enumerate}
    \item For each tree $t = 1, \ldots, T$:
    \begin{enumerate}
        \item Transform $\mathbf{x}_{new}$ into its functional representation $\mathbf{c}_{new,o_t,K_t}$ using the B-spline basis for tree $t$.
        \item Obtain the prediction $\hat{y}_t$ by traversing the $t$-th tree using $\mathbf{c}_{new,o_t,K_t}$.
    \end{enumerate}
    \item Compute the final prediction using majority voting:
    \begin{equation}
    \hat{y} = \arg\max_{z \in \{1,\ldots,Z\}} \sum_{t=1}^T \mathbb{I}(\hat{y}_t = z)
    \end{equation}
    where $\mathbb{I}(\cdot)$ is the indicator function.
\end{enumerate}

This approach ensures that each tree in the ensemble is trained and makes predictions based on a unique functional representation of the data.
The process can also be summarized by the pseudocode described in Algorithm \ref{alg:rst_inference}.

\begin{algorithm}[H]
\caption{RST Inference}\label{alg:rst_inference}
\begin{algorithmic}[1]
\Require New observation $X_{\text{new}}$, Ensemble of trees $\{T_1, T_2, ..., T_T\}$, B-spline parameters $\{\theta_1, \theta_2, ..., \theta_T\}$ 
\Ensure Predicted label $\hat{y}$

\State Initialize $vote\_count = [0, 0, ..., 0]$ \Comment{Array for votes per class}

\For{each tree $T_t$ in $\{T_1, T_2, ..., T_T\}$}
    \State Set $\theta_t = (o_t, K_t)$ \Comment{Retrieve B-spline parameters for tree $T_t$}
    \State Compute B-spline representation of $X_{\text{new}}$ using $\theta_t$: 
    \State $c_{\text{new}} \gets \text{ComputeBSplineCoefficients}(X_{\text{new}}, o_t, K_t)$
    \State Traverse tree $T_t$ with $c_{\text{new}}$ to get prediction: 
    \State $\hat{y}_t \gets T_t.\text{Predict}(c_{\text{new}})$
    \State Update $vote\_count[\hat{y}_t] \gets vote\_count[\hat{y}_t] + 1$
\EndFor

\State $\hat{y} \gets \arg\max(vote\_count)$ \Comment{Find class with maximum votes}

\State \Return $\hat{y}$ \Comment{Return final predicted label}
\end{algorithmic}
\end{algorithm}

\section{Environmental Data Analysis Using RST} \label{appl}

\subsection{Experimental setup}

This study uses a diverse set of environmental time series datasets from the UCR Time Series Archive \cite{UCRArchive}. These datasets represent various environmental phenomena and classification tasks. Table \ref{table:tsc_datasets} provides an overview of the datasets used in this study.

\begin{table}[h]
\centering
\begin{tabular}{|l|p{3cm}|c|c|c|c|c|}
\hline
\textbf{Dataset Name} & \textbf{Description} & \textbf{Train} & \textbf{Test} & \textbf{Length} & \textbf{Classes} & \textbf{Type} \\ \hline
ChlorineC & Water quality  & 467 & 3840 & 166 & 3 & Simulated \\ \hline
Rock & Rock types & 20 & 50 & 2844 & 4 & Spectrographic \\ \hline
Worms & Worm types & 181 & 77 & 900 & 5 & Motion \\ \hline
Fish & Fish species & 175 & 175 & 463 & 7 & Image \\ \hline
Earthquakes & Earthquake events & 322 & 139 & 512 & 2 & Sensor \\ \hline
ItalyPower & Power consumption patterns & 67 & 1029 & 24 & 2 & Sensor \\ \hline
\end{tabular}
\caption{Details of Time Series Classification Datasets}
\label{table:tsc_datasets}
\end{table}

The datasets were chosen to represent a range of environmental monitoring and classification tasks, including water quality assessment (ChlorineC), geological classification (Rock), biological classification (Worms, Fish), seismic activity monitoring (Earthquakes), and energy consumption patterns (ItalyPowerDemand). Each dataset comes with a predefined train-test split. The datasets include various types, from image data to motion and sensor data. 

All experiments were conducted using Python 3.12. RST's performance is compared against two of the most popular tree-based ensembles: Random Forest (RF) and Gradient Boosting (GB).

These comparison models are implemented using \texttt{scikit-learn} (version 1.5.1) \cite{scikit-learn}.

For RST, we explored several variants:
\begin{enumerate}
    \item RST-B: RST with $n_{\text{estimators}} = 100$, split strategy = \textit{best}
    \item RST-BB: RST with $n_{\text{estimators}} = 100$, split strategy = \textit{random}, bootstrap = true
    \item RST-R: RST with $n_{\text{estimators}} = 100$, split strategy = \textit{random}
    \item RST-RB: RST with $n_{\text{estimators}} = 100$, split strategy = \textit{best}, bootstrap = true
\end{enumerate}

To ensure comparability for all RST variants, the same following functional representation parameters were used.

The order of the B-spline basis $o$ for each tree was randomly selected from the range $[3, 9]$, and the number of basis functions $K$ from the range $[11,50]$. 

To ensure the comparability of our models, we maintain consistency in all hyperparameters except for the specified RST hyperparameters. This methodology ensures that any observed differences in performance can be attributed solely to the changes in the RST hyperparameters, providing a clear and controlled environment for our comparative analysis.

The provided train-test splits are used for each dataset to ensure comparability with other studies using the UCR Time Series Archive. Performance was evaluated using classification accuracy on the test set.

\subsection{Results}

This comparison aims to assess whether randomizing functional parameters benefits inducing diversity. 

\begin{table}[ht]
\centering
\begin{tabular}{lccccccc}
\hline
Dataset & GB & RF & RST-B & RST-R & RST-BB & RST-RB \\ \hline
ChlorineC & 0.7427 & 0.7156 & 0.7396 & \textbf{0.7430} & 0.7042 & 0.7094 \\
Earthquakes & 0.7482 & 0.7482 & 0.7554 & 0.7554 & 0.7482 & \textbf{0.7626} \\
Fish & 0.7029 & 0.7771 & 0.8343 & \textbf{0.8400} & 0.8114 & 0.8229 \\
ItalyPowerDemand & 0.9640 & 0.9650 & 0.9670 & 0.9689 & \textbf{0.9708} & \textbf{0.9708} \\
Rock & 0.6000 & 0.6800 & 0.6600 & \textbf{0.7400} & 0.7000 & 0.7000 \\
Worms & 0.4675 & 0.5195 & \textbf{0.5974} & 0.5714 & 0.5714 & 0.5455 \\ \hline
\end{tabular}
\caption{Comparison of different models across various datasets}
\label{tab:comparison}
\end{table}

The results of our experimental evaluation are presented in Table \ref{tab:comparison}, which compares the classification accuracy of GB, RF, and four variants of RST across six environmental time series datasets.
The RST variants, particularly RST-R (random split strategy) and RST-RB (random split with bootstrap), generally outperformed both RF and GB across the datasets. RST-R achieved the highest accuracy for three out of the six datasets (ChlorineC, Fish, and Rock), while RST-RB was the top performer for the Earthquakes dataset.
The effect of the split strategy in RST varied across datasets. For most datasets, the random split strategy (RST-R) achieved higher accuracy than the best split strategy (RST-B). However, for the Worms dataset, RST-B outperformed RST-R.
The impact of bootstrapping in RST was inconsistent across datasets. For the Earthquakes dataset, the bootstrapped variant RST-RB outperformed the non-bootstrapped RST-R. However, for datasets like ChlorineC and Fish, the non-bootstrapped variant RST-R achieved better results.
While RST variants showed superior performance on most datasets, there were exceptions. In the Worms dataset, RST-B achieved the highest accuracy among all tested models, while for the ItalyPowerDemand dataset, RST-BB and RST-RB tied for the best performance.
The magnitude of performance differences between algorithms varied across datasets. For the Fish dataset, the best-performing RST variant (RST-R) showed a 13.71\% improvement in accuracy over GB. In contrast, for the ItalyPowerDemand dataset, the performance differences were minimal, with all models achieving very high accuracy above 96\%.
Interestingly, the traditional RF and GB models were not the top performers for any of the datasets in this comparison.

\begin{figure}
    \centering
    \includegraphics[width=1\linewidth]{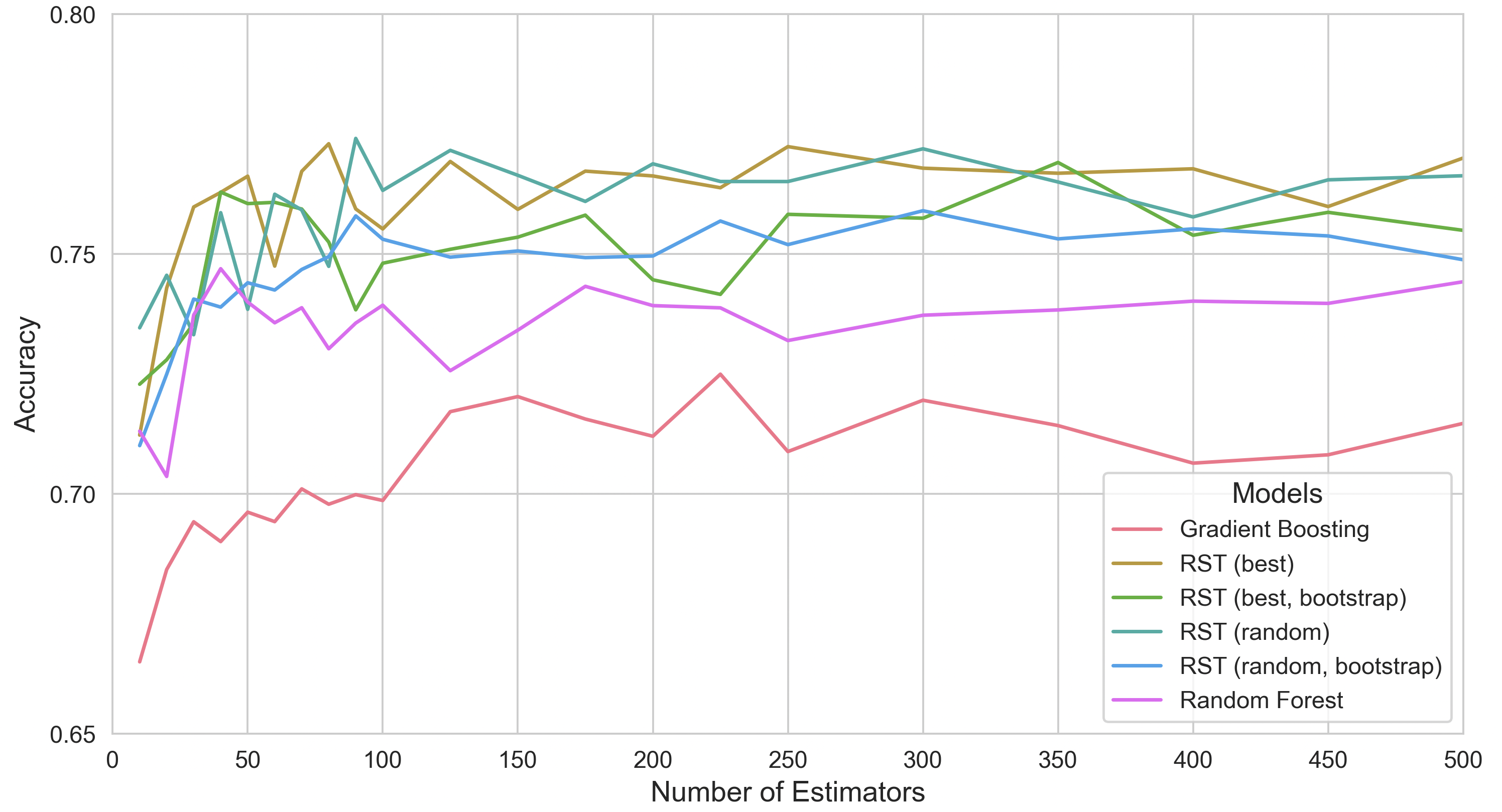}
    \caption{Average accuracy vs number of trees in the ensemble}
    \label{fig:n_e}
\end{figure}

Figure \ref{fig:n_e} illustrates the relationship between the number of estimators (trees) and the average accuracy across all datasets for different models, providing insights into the performance of these models as the ensemble size increases. All models show a general trend of performance stabilization as the number of estimators increases. The four RST variants (best, best with bootstrap, random, and random with bootstrap) consistently outperform Random Forest and Gradient Boosting across all ensemble sizes. 
The RST variants, particularly those without bootstrap, show robust performance even with few estimators. 

The RST variants using the \textit{best} split strategy generally show similar accuracy than those using the \textit{random} split strategy, indicating that the choice of this parameter has a marginal effect on performance.
On the other hand, the impact of bootstrap sampling in RST seems to impact performance negatively.
Random Forest shows steady improvement as the number of estimators increases, but its performance remains below that of the RST variants. Gradient Boosting shows a similar trend with increased estimators but fails to match RST's performance. In the 1-100 estimators range, all models show rapid improvement, with RST variants achieving high performance more quickly than traditional methods. Beyond around 200 estimators, the performance of all models appears to stabilize, with only minor fluctuations, suggesting that there may be diminishing returns in increasing the ensemble size.

\begin{figure}
    \centering
    \includegraphics[width=1\linewidth]{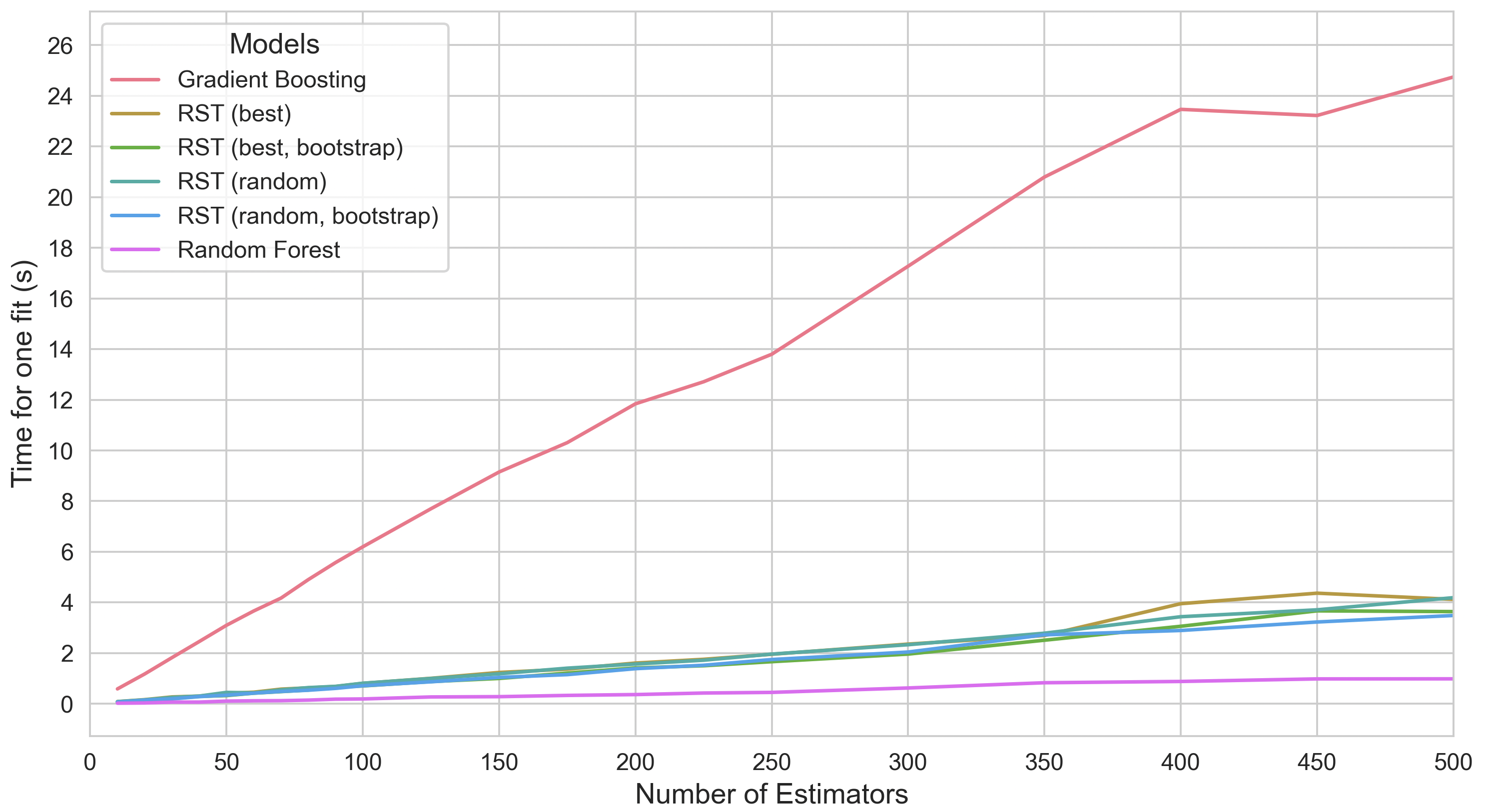}
    \caption{Time for one fit vs number of trees in the ensemble}
    \label{fig:time}
\end{figure}

Figure \ref{fig:time} illustrates the computational time required for model fitting as a function of the number of estimators for different algorithms. Gradient Boosting is the most time-intensive, RST variants are only marginally slower than Random Forest, which is the most time-efficient. All models show an increasing trend in computational time as the number of estimators increases. Gradient Boosting exhibits the steepest increase, reaching about 25 seconds for 500 estimators. The four RST variants display similar behavior, with more moderate growth in time, reaching approximately 4 seconds at 500 estimators. Random Forest shows the slowest increase, staying below 2 seconds even at 1000 estimators. The effect of bootstrap sampling and the choice of the split parameters on computational time seems minimal for RST variants.

\section{Discussion and Conclusion} \label{concl}

As environmental challenges become increasingly complex and data-driven, methods like RST that can automatically adapt to the inherent structure of environmental time series will play a key role.
The performance of RST in environmental time series classification offers several important theoretical insights.
This analysis of diversity levels in RST contributes to the broader theoretical understanding of ensemble diversity in functional data classification. The results align with the theoretical foundation of the Functional Voting Classifier \cite{riccio2024supervisedlearningensemblesdiverse}, which showed that diversity can be induced in functional voting ensembles by varying the B-spline basis representation. In our study, the randomization of B-spline parameters introduces a novel form of diversity. This functional diversity is particularly effective in capturing the complex temporal patterns inherent in environmental time series.

We explored three levels of randomization in the RST variants:
\begin{enumerate}
    \item RST-B: One level of randomization (functional representation randomization)
    \item RST-R: Two levels of randomization (functional representation randomization and random split)
    \item RST-RB: Three levels of randomization (functional representation randomization, random split, and bootstrap)
\end{enumerate}

The similar performance of RST-R and RST-B suggests that incorporating two levels of randomization is generally unnecessary. 
The functional representation randomization allows the ensemble to capture various aspects of the underlying temporal patterns, while the random splits ensure diversity in the decision boundaries. 
Adding bootstrap sampling in RST-RB did not consistently improve performance over RST-R. This suggests a potential \textit{saturation point} in the benefits of diversity for this particular application. Several factors might explain this observation. The bootstrap sampling may introduce a form of diversity that is already largely captured by the combination of functional representation randomization and random splits. While bootstrap sampling increases diversity, it also means that each tree is trained on a subset of the data. For some datasets, particularly those with limited samples, the benefits of increased diversity might be outweighed by the reduced data utilization. The optimal level of diversity may depend on the specific characteristics of the data being analyzed. Future work could explore adaptive methods that adjust the levels of diversity based on dataset characteristics.
The experimental results show competitive performance with state-of-the-art neural networks. For example, in the ItalyPowerDemand dataset, RST-BB and RST-RB match the ALSTM-FCN model in accuracy \cite{lstm}. The simplicity of the RST architecture offers significant advantages, like reduced computational cost and faster training times, making it highly efficient. Using an ensemble of tree-based models also minimizes the risk of overfitting and can be more robust in scenarios with limited data, achieving competitive accuracy with minimal training time.
These results suggest a potential trade-off between ensemble diversity and model complexity. While diversity generally improves performance, there may be a point beyond which additional sources of randomness provide minimal benefits.
RST's performance across different datasets provides empirical evidence for a nuanced bias-variance trade-off in functional spaces. The randomization of functional representations introduces an additional layer of complexity to this trade-off. While individual trees may have higher bias due to potentially suboptimal functional representations, the ensemble as a whole appears to benefit from reduced variance.
This observation suggests that the traditional understanding of the bias-variance trade-off may need to be extended when dealing with functional data. The concept of \textit{functional bias}, how well a given functional representation captures the true underlying process, becomes a key factor alongside the traditional notions of model bias and variance.
The variable performance of RST across different datasets hints at an adaptive complexity mechanism. By randomizing the number of basis functions and spline orders, RST implicitly adapts its modeling complexity to the inherent complexity of the time series. This adaptive behavior may explain RST's robustness across diverse environmental datasets with varying degrees of temporal complexity.
This finding suggests that fixed-complexity models may be suboptimal for environmental time series, which often exhibit multi-scale temporal patterns. The success of RST's adaptive approach opens up new research directions in developing models that can automatically adjust their complexity to match the data's inherent structure.

RST's approach of using randomized functional representations as a basis for classification challenges the traditional paradigm of feature engineering for time series data. Instead of manually crafting features or relying on fixed transformation techniques (e.g., Fourier or wavelet transforms), RST demonstrates the potential of learning an ensemble of representations directly from the data.
This shift from manual to learned representations aligns with broader trends in machine learning, such as the success of deep learning in image and speech recognition. However, RST achieves this in a more streamlined framework, maintaining the advantages of tree-based models.

Integrating functional data analysis (FDA) techniques with ensemble learning in RST represents a significant step towards bridging these two fields. While FDA provides powerful tools for representing and analyzing functional data, it has often been separate from mainstream machine learning research. 
This integration opens up new possibilities for developing hybrid methods that leverage the strengths of both FDA and machine learning. Future research could explore other FDA techniques (e.g., functional principal component analysis) within the RST framework or extend the RST approach to other ensemble methods.
While RST shows promising results, its computational complexity is higher than that of traditional random forests due to the functional representation step. As environmental datasets continue to grow in size and dimensionality, addressing this scalability challenge becomes crucial. Future research could explore efficient approximation methods for B-spline fitting and
parallel and distributed implementations of RST, or online learning variants of RST for streaming environmental data.

\subsection*{Declarations}

The authors declare that they received no funding for this study and have no affiliations or involvement with any organization that has a financial or non-financial interest in the subject matter of this manuscript.

\subsection*{Funding and/or Conflicts of Interest/Competing Interests}

The authors confirm that they received no support from any organization for the submitted work. They also declare no affiliations or involvement with any organization or entity that has a financial or non-financial interest in the subject matter of this manuscript.

\subsection*{Use of Generative AI in Scientific Writing}

The authors used \textit{Grammarly AI} to improve the English language in preparing this work. They reviewed and edited the content as necessary and took full responsibility for the content of the publication.

 \subsection*{Data availability statement}
 
The authors utilized publicly available data for real-world applications. Simulation data can be provided upon request.

\bibliography{sn-bibliography}
\end{document}